\documentclass[journal]{IEEEtran}
\pdfoutput=1
\usepackage{amsmath,amsfonts}
\usepackage{algorithmic}
\usepackage{algorithm}
\usepackage{array}
\usepackage{textcomp}
\usepackage{stfloats}
\usepackage{amssymb}
\usepackage{utfsym}
\usepackage{url}
\usepackage{verbatim}
\usepackage{graphicx}
\usepackage{cite}
\usepackage{amssymb}
\usepackage{multirow}
\usepackage{multicol}
\usepackage{graphicx}
\usepackage{colortbl}
\usepackage{color}
\usepackage{algorithm, algorithmic}
\usepackage[bookmarks=true, colorlinks=true, linkcolor=black, urlcolor=blue, citecolor=black,
bookmarks=true, linktocpage=true,   % makes the page number as hyperlink in table of content
hyperindex=true]{hyperref}
\begin{document}
% \bibliographystyle{ieeetr}
%\bibliographystyle{plainnat}
% \bibliography{reference}

\title{Context-Semantic Quality Awareness Network for Fine-Grained Visual Categorization}

\author{Qin~Xu, Sitong~Li, Jiahui~Wang, Bo~Jiang, Jinhui~Tang,~\IEEEmembership{Senior Member,~IEEE}
        % <-this % stops a space
% \thanks{
% This work was supported by the National Natural Science Foundation of China under Grant 72071001, by the Natural Science Foundation for the Higher Education Institutions of Anhui Province under Grant KJ2021A0038, and by the University Synergy Innovation Program of Anhui Province under Grant GXXT-2020-013 and Grant GXXT-2022-032. (\emph{Corresponding authors: Qin Xu and Bo Jiang})

% Q. Xu, S. T. Li and J. H. Wang are with the Key Laboratory of Intelligent Computing and Signal Processing of Ministry of Education, Anhui Provincial Key Laboratory of Multimodal Cognitive Computation, School of Computer Science and Technology, Anhui University, Hefei 230601, China (E-mail: xuqin@ahu.edu.cn; e22201078@stu.ahu.edu.cn; wjh@stu.ahu.edu.cn). % <-this % stops a space

% B. Jiang is with the Information Materials and Intelligent Sensing Laboratory of Anhui Province, School of Computer Science and Technology, Anhui University, and also with the Institute of Artificial Intelligence, Hefei Comprehensive National Science Center, Hefei, China. (E-mail: zeyiabc@163.com).

% J. Tang is with the School of Computer Science and Engineering, Nanjing University of Science and Technology, Nanjing, 210094, China. (E-mail: jinhuitang@njust.edu.cn)}
}


\maketitle

\begin{abstract}
Exploring and mining subtle yet distinctive features between sub-categories with similar appearances is crucial for fine-grained visual categorization (FGVC). However, less effort has been devoted to assessing the quality of extracted visual representations. Intuitively, the network may struggle to capture discriminative features from low-quality samples, which leads to a significant decline in FGVC performance. To tackle this challenge, we propose a weakly supervised Context-Semantic Quality Awareness Network (CSQA-Net) for FGVC. In this network, to model the spatial contextual relationship between rich part descriptors and global semantics for capturing more discriminative details within the object, we design a novel multi-part and multi-scale cross-attention (MPMSCA) module. Before feeding to the MPMSCA module, the part navigator is developed to address the scale confusion problems and accurately identify the local distinctive regions. Furthermore, we propose a generic multi-level semantic quality evaluation module (MLSQE) to progressively supervise and enhance hierarchical semantics from different levels of the backbone network. Finally, context-aware features from MPMSCA and semantically enhanced features from MLSQE are fed into the corresponding quality probing classifiers to evaluate their quality in real-time, thus boosting the discriminability of feature representations. Comprehensive experiments on four popular and highly competitive FGVC datasets demonstrate the superiority of the proposed CSQA-Net in comparison with the state-of-the-art methods. 
\end{abstract}


\section{Introduction}
\IEEEPARstart{F}{ine}-grained visual categorization (FGVC) aims to accurately recognize different sub-categories with near-duplicated appearances, e.g., bird species \cite{CUB}, \cite{NAB}, car brands \cite{CAR} and aircraft types \cite{maji2013fine}. FGVC has demonstrated significant practical applications across various domains, including intelligent transportation, medical diagnosis, and biodiversity conservation, thus attracting a wide range of attention. However, the inherent inter-class similarities and relatively high intra-class differences that are common in fine-grained objects pose significant challenges for this task. Therefore, representation learning of distinctive details within fine-grained objects is crucial, which has triggered a great deal of research to explore better feature representations of objects.

Based on the different ways of learning fine-grained details, the existing FGVC methods can be broadly categorized into two groups. The first category of methods focuses on encoding the high-order information to enrich feature representations \cite{bilinearcnn, cai2017higher, bilineartrans, sun2020fine, zhuang2020learning, ding2021ap,TOAN} or imposing auxiliary constraints \cite{sun2018multi, CNENet}. For example, Sun \emph{et al.} \cite{sun2018multi} suggest enhancing the spatial correlation between different image regions by multi-attention multi-constraint (MAMC). However, this group of methods faces two main challenges: 1) They merely encode the semantic features of objects from a global perspective, neglecting discriminative regions like local part features, which are crucial for capturing the subtle appearance differences between sub-categories; 2) The high-level semantics extracted from deep layers only contain the abstract and structural information, while the low-level detailed clues (e.g., texture or contoured edge) in relatively shallow layers are overlooked.

To effectively explore the distinctive part features, the other group of FGVC methods locates the class-discriminative regions by localization sub-modules or attention mechanisms. Early pioneering works \cite{krause2015fine, Part-stacked, spda-cnn} identify salient parts by introducing object bounding boxes or fine-grained level part annotations. Nevertheless, such labor-intensive annotations are costly and demand professional expertise. Thus, the weakly supervised FGVC methods \cite{he2019fast, navigate, crosspart, fad, 9614988, P-CNN, CAP}, which rely solely on image labels to discover the key object parts, are proposed. For example, Behera \emph{et al.} \cite{CAP} propose a novel attention mechanism, named context-aware attention pooling (CAP), which is related to the sub-pixel gradients and focuses on the importance of informative regions and encodes the semantic relationships between different parts. Although previous weakly supervised methods have achieved impressive performance, they rarely consider or only model the correlation between distinctive parts at a single-scale level. Furthermore, they fail to establish an effective learning mechanism between global semantics and local parts. These limitations hinder their ability to capture more detailed texture and structural information, thereby restricting their performance in identifying fine-grained images with near-duplicated appearance.
\begin{figure}[!t]
\centering
\includegraphics[width=1\linewidth]{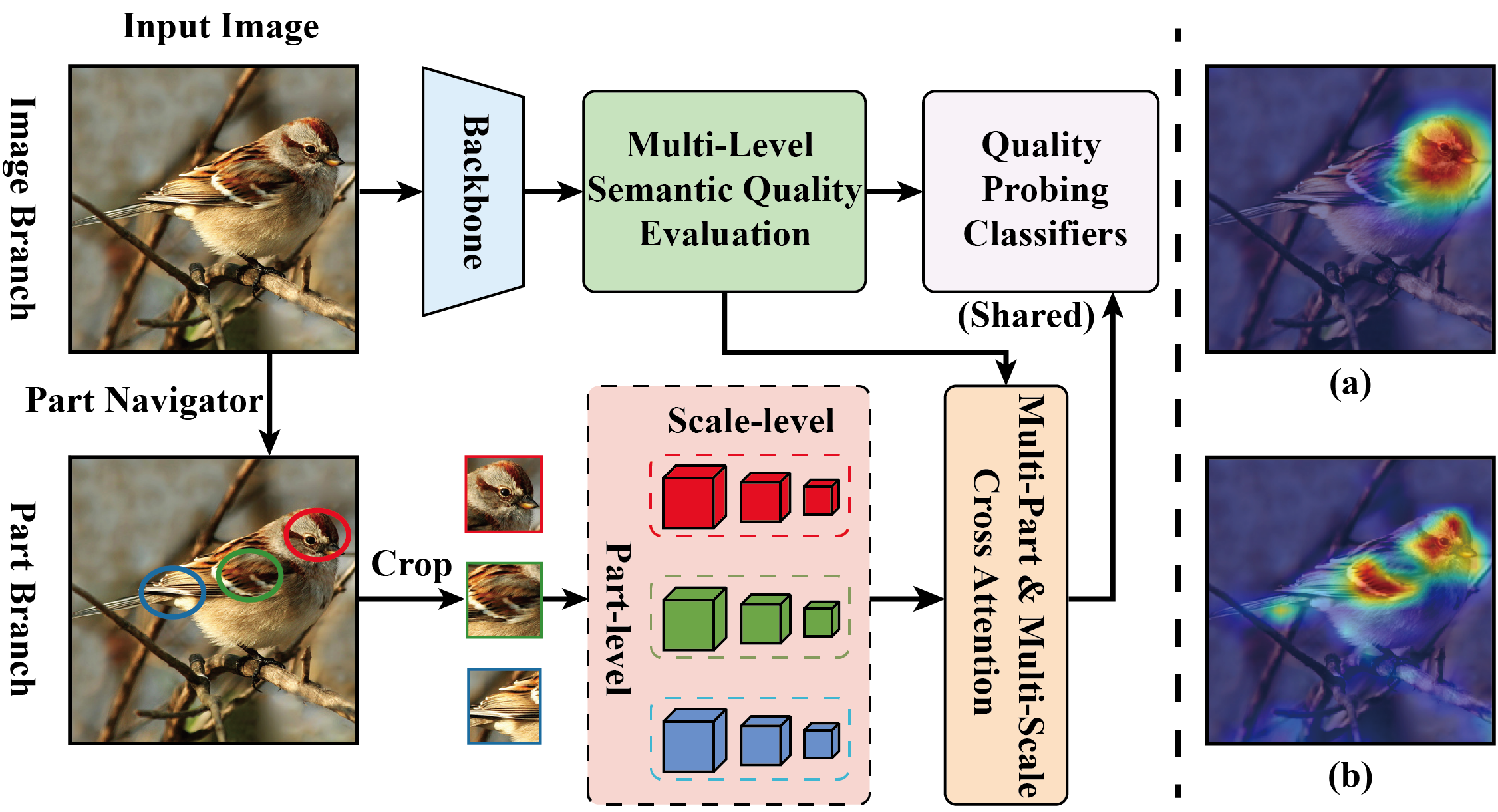}
\caption{A brief view of CSQA-Net. For the image branch, we use multi-level semantic quality evaluation module to enhance hierarchical semantics extracted from the backbone. For the part branch, part navigator is utilized to locate the discriminative regions, and multi-part and multi-scale cross-attention is proposed to generate context-aware features. Finally, we use quality probing classifiers to assess and improve the quality of visual representations. (a) and (b) represent the key response regions generated by ResNet-50 and our proposed CSQA-Net, respectively. }
\label{fig: motivation}
\end{figure}

To efficiently tackle the aforementioned problems within a unified framework, we propose an end-to-end \textbf{C}ontext-\textbf{S}emantic \textbf{Q}uality \textbf{A}wareness \textbf{Net}work (CSQA-Net). Different from the existing FGVC methods, which integrate the hierarchical features from shallow to deep on the image level, we focus on the interaction of multi-scale part features with the global semantics, thus simulating contextual relationships between objects and parts and capturing subtle yet discriminative differences. Specifically, we propose a \textbf{M}ulti-\textbf{P}art and \textbf{M}ulti-\textbf{S}cale \textbf{C}ross-\textbf{A}ttention (MPMSCA) module. Diverging from the vanilla self-attention and cross-attention mechanisms, the MPMSCA module integrates global objects and local parts into query vectors from a multi-scale perspective, then multiplies these query vectors with the key value vectors composed of rich part descriptors for spatial context interaction. Therefore, this MPMSCA that models long-range dependencies is beneficial to alleviate the impact of pose and scale changes within fine-grained objects, thus eliminating redundant noise and automatically mining discriminative clues, for providing context-aware robust features.

Moreover, most of the existing FGVC works merely consider well-designed network structures or modules to encode feature representations, while few efforts have been devoted to assessing the quality of these visual representations. We consider that real-time evaluation of semantic quality across different levels is essential for FGVC and will provide a novel thinking perspective for regularized representation learning. To be specific, we propose a \textbf{M}ulti-\textbf{L}evel \textbf{S}emantic \textbf{Q}uality \textbf{E}valuation (MLSQE) module that progressively supervises and fully exploits the complementary information from feature maps generated at different stages, in which \textbf{Q}uality \textbf{P}robing (QP) Classifiers are designed to evaluate the linear separability of visual representations in real-time, thus prompting their discriminability. It is worth noting that QP classifiers are weight-sharing, that is, the final output vectors of global semantics and local parts share the corresponding QP classifiers, and the MLSQE module can be easily combined with the hierarchical backbone and stably improve FGVC performance. 

Benefiting from the proposed MPMSCA and the MLSQE modules, our CSQA-Net can discover and recover the subtle yet distinctive clues buried in object representation that were usually ignored by most works, as presented in Fig. \ref{fig: motivation}. Notably, the part navigator and the MPMSCA module are only activated during the training phase. Therefore, our CSQA-Net does not rely on the part branch during the testing phase to improve computational efficiency.

In summary, our major contributions are as follows: 
\begin{itemize}
    \item We propose an end-to-end Context-Semantic Quality Awareness Network (CSQA-Net), which explores more detailed part descriptors to regularize global semantics and improves the quality of visual representation through real-time evaluation, thus mining inactivated distinctive clues buried in global features.
    \item We develop a generic multi-level semantic quality evaluation module to progressively supervise shallow-to-deep semantic information, where novel quality probing classifiers encourage the features to be more generalized and discriminated, providing better representation learning.
    \item We propose part navigator and multi-part and multi-scale cross-attention for the part branch: 1) Part navigator for alleviating scale confusion problems and locating discriminative parts; 2) Multi-part and multi-scale cross-attention for exploring visual-spatial relationships between objects and parts from multi-scale perspective to enhance the context-awareness of features.
    \item Comprehensive experiments on four popular and highly competitive benchmarks (i.e., CUB-200-2011, Stanford Cars, FGVC-Aircraft, and NABirds) verify the superior performance of the designed CSQA-Net.
\end{itemize}

% The rest of this paper is organized as follows: We briefly summarize some previous related works in Section \ref{sec:related work} and describe the proposed CSQA-Net in Section \ref{sec:approach}. Then in Section \ref{sec:experiment}, we discuss the experimental results in detail. Finally, the full paper conclusion is presented in Section \ref{sec:conclude}.

\section{Related Work}
\label{sec:related work}
In this section, we briefly review works closely related to our proposed method, i.e., feature representation learning, part-based methods, and attention-based methods for FGVC.

\subsection{Feature Representation Learning}
Deep Convolutional Neural Networks (CNNs) \cite{NIPS2012_c399862d,szegedy2015going,he2016deep} have been widely studied, but they often struggle to effectively extract the discriminative features necessary for fine-grained recognition. To this end, researchers have proposed well-designed high-order feature encoding methods to learn improved representations. For example, Lin \emph{et al.} \cite{bilinearcnn} propose bilinear pooling to encode the high-order representation of images through dual-stream CNN. Based on this, Zheng \emph{et al.} \cite{bilineartrans} design bilinear transformation operation grouped by channels to reduce the high computational cost caused by interactive learning. In \cite{sun2020fine}, a suppression module that masks high-response regions is proposed to force the network to focus on different categories that only contain subtle differences. Encouraged by the feature pyramid, Ding \emph{et al.} \cite{ding2021ap} design a pyramid structure of feature and attention paths in opposite directions to enhance the region of interest.

% Recently, Zhuang \emph{et al.} \cite{zhuang2020learning} propose a pairwise interactive attention network, which learns to guide the network to capture discriminative cues through interactive learning between image pairs.

% \citet{wang2023semantic} proposed a semantic-driven alignment network that can dynamically select shallow detail information and perform complementary learning with deep information. 

However, these methods solely focus on encoding deep features but ignore the complementary information between features at different levels. To solve this problem, Du \emph{et al.} \cite{PMG} introduce the progressive training strategy to the FGVC task for the first time to fully utilize multi-granularity information. Although PMG \cite{PMG} achieves good performance, the huge training time makes it unfavorable for large-scale fine-grained recognition tasks. To solve this problem, Yang \emph{et al.} \cite{yang2022fine} regularize the global features and part information through a self-supervised pose alignment method, and propose curriculum supervision to promote distinctive features. Unlike the above methods, we measure and enhance the feature's quality at different levels in real-time through quality probing classifier, and progressively guide the network to learn complementary information at different granularities, thus emphasizing discriminative clues and providing robust representations.

\subsection{Part-Based Methods}
Another effective way to address the FGVC task involves locating distinctive parts. In view of this, many part-based methods have been proposed to capture the subtle differences within fine-grained images. Early works rely on part annotations or bounding boxes to discover part clues. For example, Krause \emph{et al.} \cite{krause2015fine} propose to combine co-segmentation and alignment to generate target parts, which uses bounding boxes to assist training and testing. Huang \emph{et al.} \cite{Part-stacked} introduce a part detection sub-network into the framework and simultaneously extract the features of detected parts and objects in a dual-stream network. Moreover, Zhang \emph{et al.} \cite{spda-cnn} construct a detection and classification sub-network to jointly perform part extraction and identification. Although additional annotations are advantageous for fine-grained recognition, obtaining them requires expertise and is time-consuming.

Consequently, weakly supervised part localization methods using only category labels have gradually become mainstream methods. \cite{navigate,crosspart} use well-designed self-supervised region proposal networks to locate informative regions. Furthermore, Liu \emph{et al.} \cite{fad} propose novel performance metrics for region proposal networks and utilize filtering methods to select the most discriminative regions. Zhao \emph{et al.} \cite{9614988} first discover distinctive parts, and then use relational transformation to individually construct relational embeddings of global semantics or local part features to advance FGVC performance. Han \emph{et al.} \cite{P-CNN} perform final classification on the located parts individually or combined with global features, and design multiple loss functions for joint optimization. However, most of these works merely focus on obtaining local information regions, while ignoring the establishment of effective interactive learning mechanisms between detected salient regions. In contrast, we introduce a novel mutual learning strategy aimed at modeling the spatial contextual relationships between object-parts and part-part, to mine subtle yet differentiated clues between similar sub-categories.

\subsection{Attention-Based Methods}
In addition to the methods mentioned above, some works also incorporate attention mechanisms to detect and learn discriminative features in FGVC. For example, Ding \emph{et al.} \cite{ding2019selective} identify objects and discriminative parts by selectively paying attention to the maximum value of the learned class response map via sparse attention. Liu \emph{et al.} \cite{8907499} propose a bidirectionally enhanced attention agent and a recognition agent to optimize the filtered-out regions. Furthermore, Ji \emph{et al.} \cite{ji2020attention} design a binary neural tree that combines convolution and attention operations, and calculates the final prediction based on the root node. Rao \emph{et al.} \cite{CAL} propose counterfactual attention strategies based on causal relationships to improve the network's ability to learn higher-quality attention. 

More recently, vision Transformers, which rely on the self-attention mechanism as their core component, have been introduced into computer vision tasks. As a pioneering work, ViT \cite{dosovitskiy2020image} splits images into a series of image patches with positional embeddings and uses a vanilla Transformer architecture to achieve amazing performance in various vision tasks. However, ViT \cite{dosovitskiy2020image} is proficient in modeling long-range dependencies but lacks the ability to locate distinctive parts, which results in its advantages in FGVC tasks not being fully reflected. To solve this problem, He \emph{et al.} \cite{he2022transfg} propose to select tokens with higher response values based on the attention map of the previous layer and input them to the last layer. Liu \emph{et al.} \cite{liu2021swin} propose a hierarchical Swin Transformer, which takes into account both computational complexity and scale flexibility through shifted windows. Ji \emph{et al.} \cite{Dual-TR} design parallel global and local Transformer architectures to perform orthogonal assembly in two branches to obtain more robust features. Different from the attention mechanism used above, we propose a novel cross-attention operation, which reorganizes local and global information into query vectors and performs cross-attention with key value vectors of multi-scale parts, thus learning more discriminative and robust feature embeddings for fine-grained objects identification.

\begin{figure*}
    \centering
    \includegraphics[width=1\linewidth]{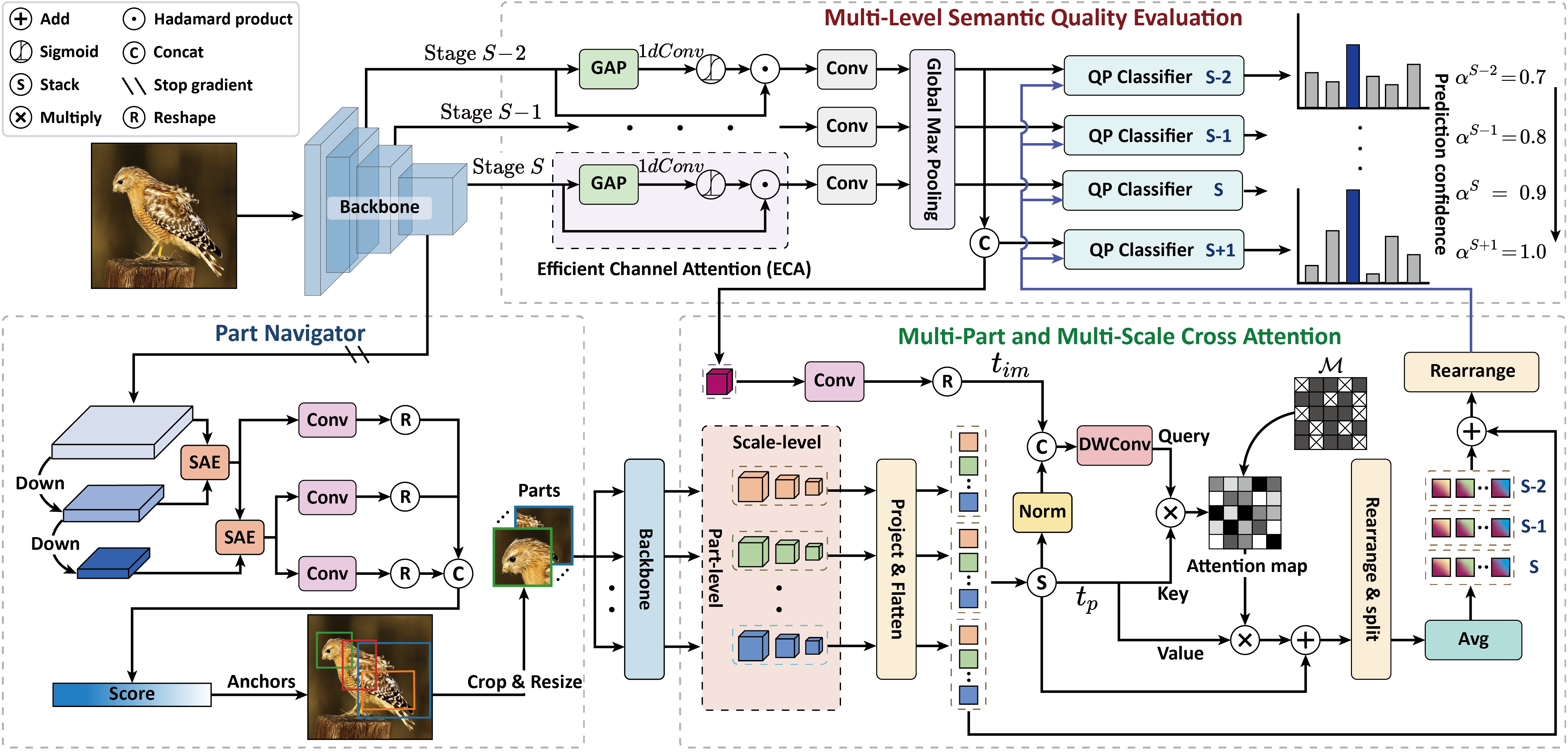}
    \caption{The detailed architecture of CSQA-Net, which consists of feature extractor (Backbone parameters in light blue and black font are shared), multi-level semantic quality evaluation module, part navigator, and multi-part and multi-scale cross-attention module. $S$ denotes the number of stages included in the backbone network. For clarity, we set $A$ to 3. $\alpha$ represents the confidence for the output of different stages. The detailed structure of the quality probing classifier (QP Classifier) and scale-aware enhancement (SAE) block are shown in Fig. \ref{fig: quality probing classifier} and Fig. \ref{fig:part navigator}, respectively. }
    \label{fig:network}
\end{figure*}

\section{Approach}
\label{sec:approach}
The network architecture of the proposed CSQA-Net is presented in Fig. \ref{fig:network}. Technically, our framework mainly consists of three modules: 1) Multi-level semantic quality evaluation uses multi-stage progressive learning strategy to measure and enhance the quality of semantic features at different levels in real time; 2) A part navigator is designed to mine local regions of interest in fine-grained objects, which reduces the impact of redundant background environments on classification; 3) With the detected salient regions, we propose a multi-part and multi-scale cross-attention to perform feature interactions between detailed part descriptors and global object representations, thus regularizing object representation learning.

\subsection{Multi-Level Semantic Quality Evaluation}
\label{sec:progressive}
In classifying the fine-grained images, samples of varying difficulty levels exhibit different quality levels, and some features often lead to incorrect classification results, indicating their lower quality. Hence, it is necessary to evaluate and enhance feature quality during the training phase to improve recognition performance. However, the existing methods rarely consider this aspect. In addition, judiciously incorporating low-level detailed information to supplement deep abstract semantics can effectively mitigate intra-class variations. Based on the above analysis, we propose the multi-level semantic quality evaluation module (MLSQE), which progressively supervises and evaluates the quality of visual representations at different stages in an online manner. 

As Fig. \ref{fig:network} shows, given the input image, ResNet-50 \cite{he2016deep} or Swin Transformer \cite{liu2021swin} is adopted as the backbone, which contains $S$ stages. Here, the output feature maps from the last $A$ stages are used, and the feature map of the $s$-th stage can be expressed as $F_{im}^s \in \mathbb {R}^{H^s\times W^s\times C^s}$, where $s\in \left\{S \! - \! A\! + \!1, S\! - \!A\! + \!2,\cdots, S\right\}$, and $H^s$, $W^s$, $C^s$ respectively represents the height, width and channel dimensions. As different feature map channels often contain different visual clues (e.g., the head or tail feathers of a bird), we use the lightweight ECA \cite{wang2020eca} to perform the interaction between channel information to improve the FGVC performance. The calculation is as follows: 
\begin{equation}
    \hat F_{im}^s = \sigma \left( {\kappa \left( {{\rm{GAP}}\left( {F_{im}^s} \right)} \right)} \right) \odot F_{im}^s
\end{equation}
where $\sigma$ is a sigmoid function, $\kappa$ and GAP$(\cdot )$ denote one-dimensional convolution and global average pooling respectively, and $\odot$ represents the Hadamard product.

Then the enhanced feature $\hat F_{im}^s$ is input into the convolution block $B^s$ and global maximum pooling (GMP), projecting the channel dimension to $C^*$ and compressing the spatial dimension to obtain the feature vector $v_{im}^s$, where $C^*$ is a hyper-parameter. To fully take advantage of feature fusion, we concatenate the feature vectors from the last $A$ stages along the channel dimension to get a multi-scale fusion vector ${v_{im}^{S+1}} = \left[ {v_{im}^{S- A+1}:v_{im}^{S - A + 2}:\cdots:v_{im}^S} \right]$. Finally, to generate more discriminative and generalizable features, we feed $v_{im}^{S - A+1}$, $v_{im}^{S - A+2}$, $\cdots$, $v_{im}^{S}$ and ${v_{im}^{S+1}}$ to the corresponding quality probing (QP) classifier respectively for classification. 

% ${v_{im}^{A+1}} = \left[ {v_{im}^{A-S+1}:v_{im}^{A-S+2}:v_{im}^A} \right]$. Finally, to generate more discriminative and generalizable features, we feed $v_{im}^{A-S+1}$, $v_{im}^{A-S+2}$, $v_{im}^{A}$ and ${v_{im}^{A+1}}$ to the corresponding quality probing (QP) classifier respectively for classification. 

\begin{figure}[!t]
\centering
\includegraphics[width=1\linewidth]{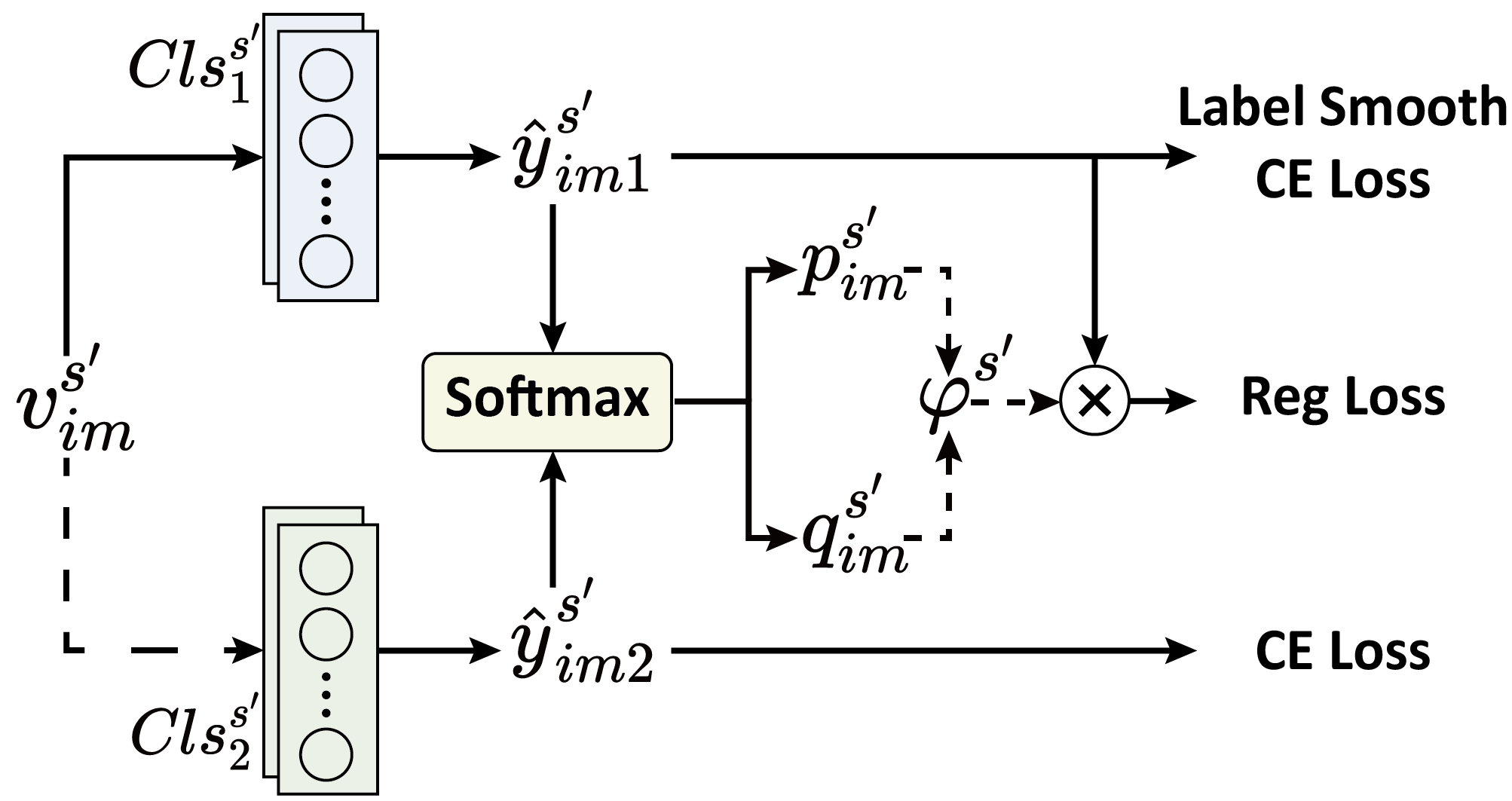}
\caption{Illustration of quality probing (QP) classifier. Solid and dotted lines indicate with and without gradient back-propagation, respectively.}
\label{fig: quality probing classifier}
\end{figure}

For real-time evaluation of different learned semantic features, inspired by \cite{liang2022simple}, a novel quality probing (QP) classifier is developed, as shown in Fig. \ref{fig: quality probing classifier}. Concretely, the vector $v^{s'}_{im}$ is fed to the main classifier $Cls_1^{s'}$ and the auxiliary classifier $Cls_2^{s'}$ respectively to obtain the corresponding prediction results $ \hat y_{im1}^{s'}$ and $\hat y_{im2}^{s'}$, where $s'\in \left\{S \! - \! A\! + \!1, S\! - \!A\! + \!2,\cdots, S+1\right\}$, then the label smoothing cross entropy loss is used to constraint $\hat y_{im1}^{s'}$ with the smoothing factor ${\alpha ^{s'}}$, where ${\alpha^{s'} \in} \ \left\{{0,\cdots,1} \right\}$. The loss function can be expressed as follows:
\begin{equation}
\begin{aligned}
   \ell _{sce}(\hat y_{im1}^{s'},y_{\alpha ^{s'}})&=\sum\limits_{i = 1}^C { - {y_{{\alpha ^{s'}}}}\left[ i \right]\log \left( {\hat y_{im1}^{s'}\left[ i \right]} \right)} \\
{y_{{\alpha ^{s'}}}}[i] &= \left\{ {\begin{array}{*{20}{l}}
{{\alpha ^{s'}}}&{i = y,}\\
{\frac{{1 - {\alpha ^{s'}}}}{C}}&{i \ne y.}
\end{array}} \right.
\end{aligned}
\end{equation}
where $l_{sce}$$(\cdot )$ represents the label smoothing cross entropy loss, $y_{\alpha^{s'}}$ is the changed ground truth label, $i$ denotes the element index of the label vector $y \in \mathbb {R}^C$ and $C$ is the number of classes. Notably, ${\alpha ^{s'}}$ signifies the confidence level of the changed ground truth label. Intuitively, deeper layers of the neural network possess better representation learning capabilities, so we gradually increase the value of ${\alpha ^{s'}}$, as shown on the right side of Fig. \ref{fig:network}. As ${\alpha ^{s'}}$ increases, $y_{\alpha^{s'}}$ gets closer and closer to the one-hot label, which encourages deeper sub-networks to make more confident predictions.

After that, the standard cross entropy loss is utilized on $\hat y_{im2}^{s'}$ to train $Cls_2^{s'}$ for correctly classifying features into $C$ classes, which is defined as follows:
\begin{equation}
  \ell _{ce}(\hat y_{im2}^{s'},y)=\sum\limits_{j = 1}^C { - {y[j]}\log \left( {\hat y_{im2}^{s'}} [j]\right)}
\end{equation}
where if $j$ is the ground truth label, $y[j]$ = 1, otherwise, $y[j]=0$. Note that, $Cls_2^{s'}$ does not propagate gradients back to the network, ensuring that it remains unaffected by the main classifier and can objectively assess the discriminability of the feature representation. In order to prevent $Cls_2^{s'}$ from over-fitting, we re-initialize its parameters $W$ and $b$ every $\delta$ epochs, where $W$ represents the weight matrix, $b$ is the bias, and $\delta$ is a hyper-parameter.

To effectively regularize network representations using the auxiliary classifier, it is crucial to construct a suitable regularization term. Therefore, we construct a factor $\varphi^{s'}$ for each sample in the $s$-th stage to dynamically adjust the label smoothing cross entropy loss: 
\begin{equation}
\ell _{reg^{im}}^{s'}= {\mathop{\rm detach}\nolimits}(\varphi^{s'})*\ell _{sce}(\hat y_{ im1}^{s'},y_{\alpha ^{s'}})
\end{equation}
where detach$(\varphi^{s'})$ means that we detach the gradient from $\varphi^{s'}$ so that this factor merely affects the size of the loss gradient. If a sample is less discriminative, in other words, the sample is more challenging to classify correctly, then $\varphi^{s'}$ will impose a relatively larger value, prompting the network to pay more attention to this sample.

More specifically, we pass $\hat y_{im1}^{s'}$ and $\hat y_{im2}^{s'}$ through softmax and max functions respectively to get the maximum probability score $p^{s'}_{im}$ and $q^{s'}_{im}$. These scores are then utilized to construct the adaptive factor $\varphi^{s'}$. First, we evaluate the distance between $p^{s'}_{im}$ and $q^{s'}_{im}$ because it can measure the confidence gap between the two classifiers. If the distance is small, it means that the sample has relatively strong discriminability. Moreover, we should also consider the sum of the two scores. If both scores are low, it indicates that the network has not learned the sample well, so we construct $\varphi^{s'}$ as follows:
\begin{equation}
    \varphi^{s'}= \frac {\left|p^{s'}_{im}-q^{s'}_{im} \right|} {p^{s'}_{im}+q^{s'}_{im}}
\end{equation}

\begin{figure}[!t]
\centering
\includegraphics[width=1\linewidth]{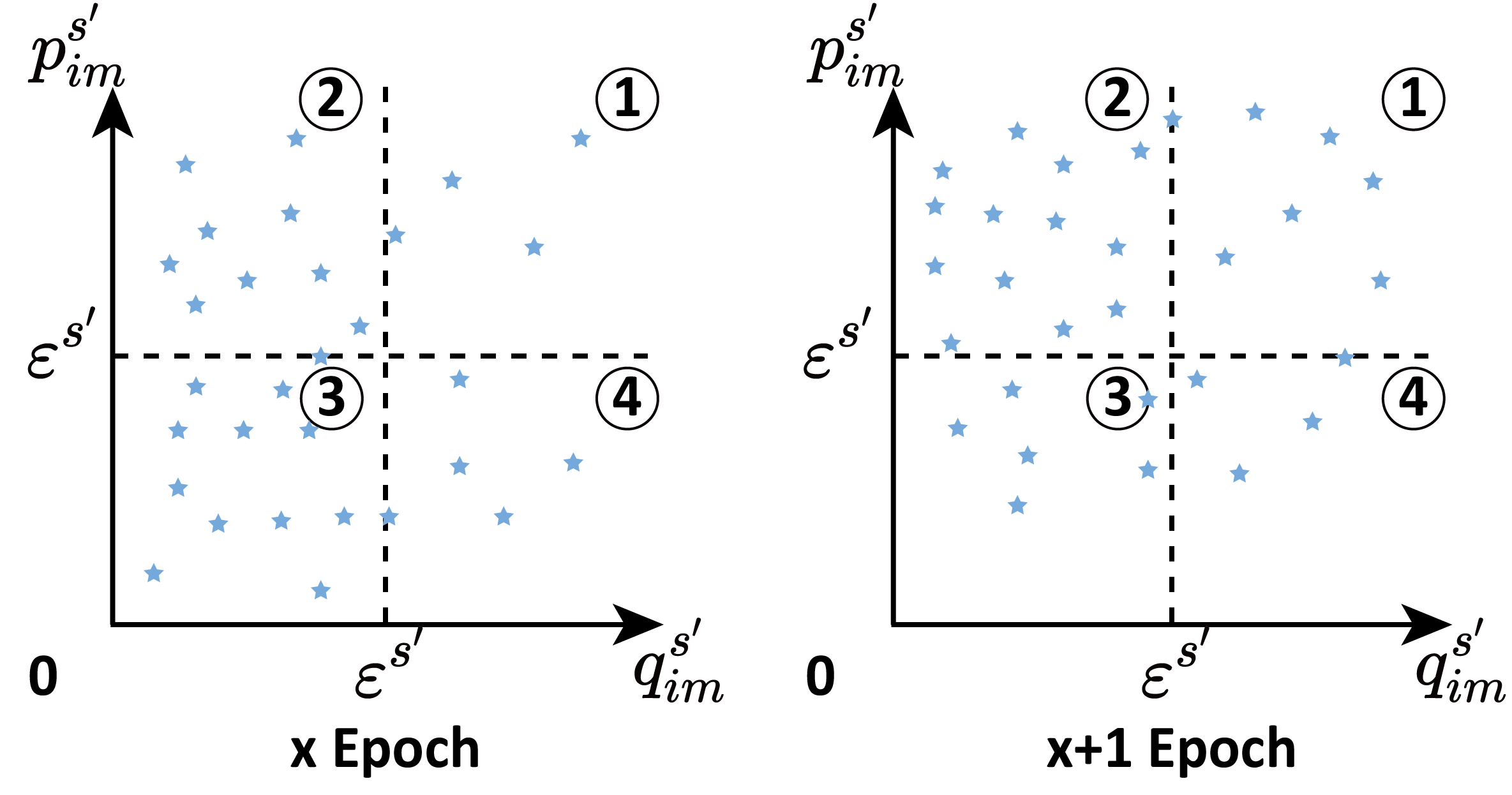}
\caption{Illustration of dividing the regions according to $\varepsilon^{s'}$. $x$ and $x$+1 represent two adjacent epochs in the training phase.}
\label{fig: distribution}
\end{figure}

Notably, we can distribute the samples in different regions according to $p^{s'}_{im}$ and $q^{s'}_{im}$, as depicted in Fig. \ref{fig: distribution}, where $\varepsilon^{s'}$ is a hyper-parameter related to ${\alpha}^{s'}$. Intuitively, $\varphi^{s'}$ is always less than 1, and it is inappropriate to use the same loss function to constrain samples with different scores. For samples in region 1, we should impose a smaller $\varphi^{s'}$, as the feature points in this area can be considered approximately linearly separable, that is, they are of higher quality. In contrast, for the sample points in regions 2, 3 and 4, we should impose a larger $\varphi^{s'}$, but the samples in region 3 need more attention because neither classifier accurately classified them. Therefore we further impose an adjustment coefficient $\lambda$:
\begin{equation}
{\varphi ^{s'}} = \left\{ {\begin{array}{*{20}{l}}
{{{({\varphi ^{s'}})}^\lambda }}&{{\rm{if}}\;(p_{im}^{s'},q_{im}^{s'})\;{\rm{in}}\;{\rm{region}}\;1,}\\
{{\varphi ^{s'}} \times \frac{\lambda }{2}}&{{\rm{if}}\;(p_{im}^{s'},q_{im}^{s'})\;{\rm{in}}\;{\rm{region}}\;2\;{\rm{or}}\;4,}\\
{{\varphi ^{s'}} \times \lambda }&{{\rm{otherwise}}.}
\end{array}} \right.
\end{equation}
where $\lambda$ smoothly adjusts the intensity of $\varphi^{s'}$. As Fig. \ref{fig: distribution} shows, the goal of QP classifiers is to efficiently transfer sample points from lower scoring regions (e.g., region 3) to higher scoring regions (e.g. regions 1 and 2) between two adjacent epochs, thus providing better visual representations.

In summary, the total classification loss for each image can be expressed as:
\begin{equation}
\label{equ:loss}
 {{\mathcal L}_{im}} = \sum\limits_{s' = S-A+1}^{S + 1} {\left[ {{\ell _{sce}}(\hat y_{im1}^{s'},{y_{{\alpha ^{s'}}}}) + {\ell _{ce}}(\hat y_{im2}^{s'},y) + \ell _{re{g^{im}}}^{s'}} \right]} 
\end{equation}
\subsection{Part Navigator }
The learning of discriminative part features helps identify subtle differences within fine-grained images. Recently, some FGVC methods \cite{navigate}, \cite{yang2022fine} adopt the pyramidal feature hierarchy to locate the discriminative parts. However, there exists a problem of feature scale confusion in this architecture. Specifically, deep features with rich semantic information are conducive to the detection of large-scale components, while shallow features, which capture detail and texture information, are better suited for detecting small-scale components. Nevertheless, large-scale component information is mixed in the shallow layers, and with the increase of convolutional layers, the network gradually emphasizes large-scale components while diminishing small-scale component information. Motivated by this, to accurately detect discriminative parts across different scales, we propose the Part Navigator.

We build the Part Navigator to process the output feature map $F_{im}^S$ of the backbone. First, we use consecutive 3×3 convolutions to generate feature maps in descending order of scale, denoted as $\left\{ {{f_1},{f_2},\cdots,{f_M}} \right\}$, where $M$ is a hyper-parameter. After that, to alleviate the issue of feature scale confusion, we design the scale-aware enhancement (SAE) block and apply it to the adjacent two layers of feature maps with different scales. Fig. \ref{fig:part navigator} shows the details of SAE. In the SAE block, assuming the input feature maps are $f_m$ and $f_{m+1}$, the process is as follows:
\begin{figure}[!t]
    \centering
    \includegraphics[width=1\linewidth]{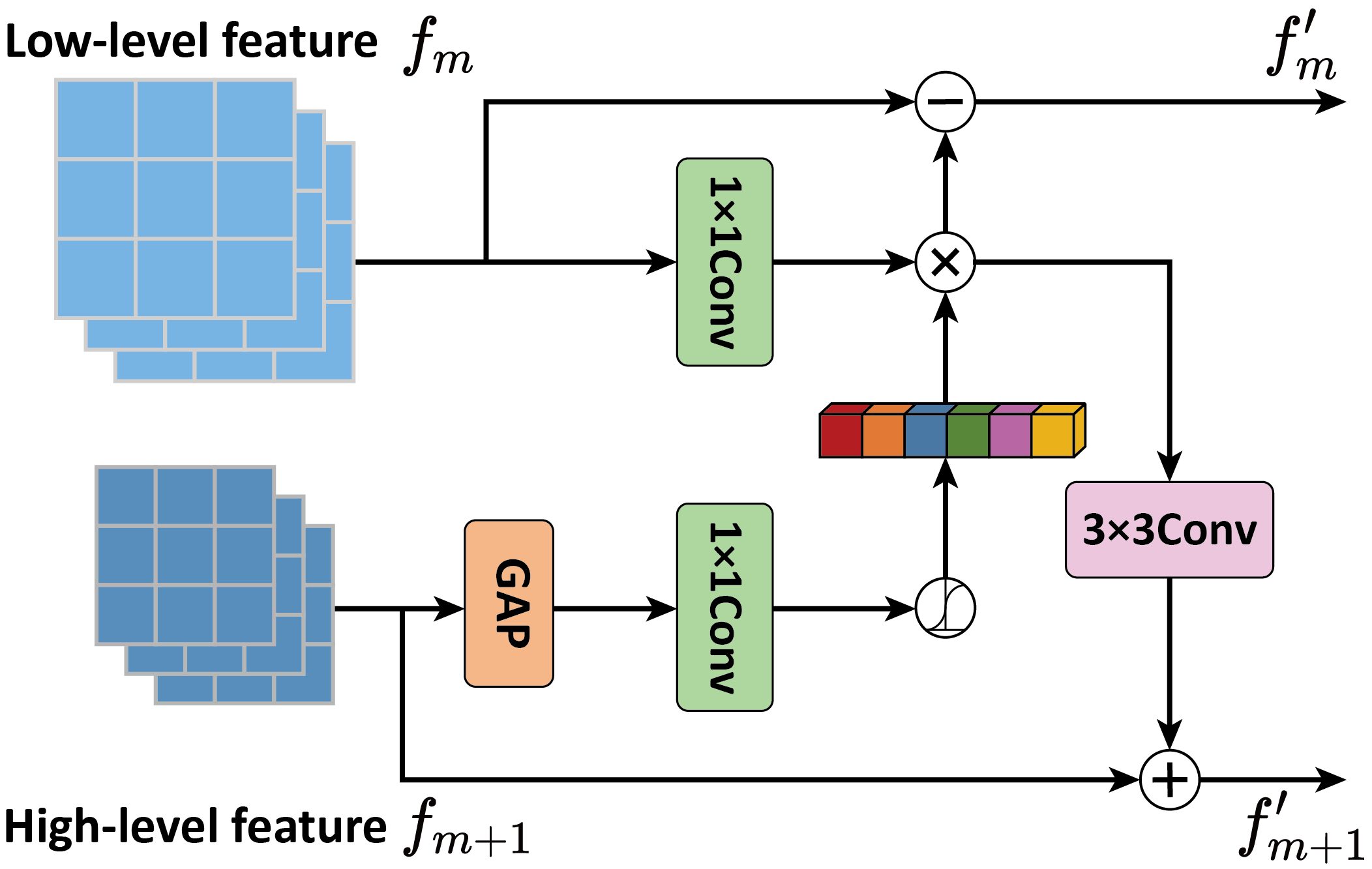}
    \caption{Illustration of scale-aware enhancement (SAE) block, which is embedded in part navigator to alleviate the scale confusion problem.}
    \label{fig:part navigator}
\end{figure}
\begin{equation}
    \begin{aligned}
        {f'_m} =&f_m \ominus \left( {\Omega \left( {{f_m}} \right) \otimes \sigma \left( {\Omega \left( {\mathop{\rm GAP}\left( f_{m+1} \right)} \right)} \right)} \right) \\[1.5mm]
    {f'_{m+1}}=f&_{m+1} \oplus \Psi \left( {\Omega \left( {{f_m}} \right) \otimes \sigma \left( {\Omega \left( {\mathop{\rm GAP}\left( f_{m+1} \right)} \right)} \right)} \right)
    \end{aligned}
\end{equation}
where $\Omega$, $\Psi$ respectively represents 1×1 convolution and 3×3 convolution. $\ominus$, $\otimes$, $\oplus$ denotes element-wise subtraction, element-wise multiplication, and element-wise addition, respectively, and $\sigma$ is a sigmoid function. The designed SAE is used for hierarchical computation on the adjacent feature maps. First, $f_1$ and $f_2$ are fed to the SAE block, and get $f'_1$ and $f'_2$. Then $f'_2$ and $f_3$ are processed by SAE to get $f''_2$ and $f'_3$. Similarly, a series of transformed feature maps are generated and get the final output $\mathcal{T}=\left\{ f'_1,f''_2,f''_3,\cdots,f'_M \right\}$. 

To better balance the trade-off between efficient computation and accuracy, we set $M$ = 3 in our experiments.  For each transformed feature, 1$\times$1 convolution is utilized to generate feature maps with different spatial resolutions, such as $\left\{ {14 \times 14,7 \times 7,4 \times 4} \right\}$. Subsequently, we reshape and concatenate them into a vector $I$ representing the scores. Following the settings in \cite{navigate}, we apply three scale anchors $\left\{ {48,96,192} \right\}$ to locate more subtle discriminative regions. Note that, the vector $I$ reflects the probability that informative regions exist in anchors. To eliminate redundant regions, we employ non-maximum suppression (NMS) to filter out the top $N$ bounding boxes based on $I$, where $N$ is a hyper-parameter representing the number of parts. Finally, we crop image patches from the input image according to the indices of the top $N$ bounding boxes and resize them to half the spatial resolution of the original input image for efficient computation. In brief, the Part Navigator mines $N$ discriminative parts, and Algorithm~\ref{alg:algorithm1} summarizes its main process. Moreover, the proposed part navigator can be trained end-to-end without requiring additional annotations, which has a unique advantage compared with other localization methods.

\begin{algorithm}
	\renewcommand{\algorithmicrequire}{\textbf{Input:}}
	\renewcommand{\algorithmicensure}{\textbf{Output:}}
	\caption{Part Navigator}
    \label{alg:algorithm1}
	\begin{algorithmic}[1]
		\REQUIRE Multi-scale features $\left\{ {{f_1},{f_2},\cdots,{f_M}} \right\}$, Number of parts $N$, IoU threshold $\theta$, default anchors $A$
		\ENSURE Part Set: $\mathcal{P}  = \left\{ {{p_1},{p_2},...,{p_N}} \right\}$
		\STATE Initialize the transformed features $\mathcal{T}=\varnothing$ 
  \FOR{$m=1$ to $M\!-\!1$}
        \STATE $\left( {{f'_m},{f'_{m + 1}}} \right) = \mathop{\rm SAE} \left( {{f_m},{f_{m + 1}}} \right)$
        \IF {$m = M\!-\!1$}
         \STATE $\mathcal{T} = \mathcal{T} \cup \left\{ f'_m,f'_{m + 1} \right\}$
        \ELSE
        \STATE Update $f_{m + 1}$ with $f'_{m + 1}$
        \STATE $\mathcal{T}=\mathcal{T}\cup \left\{f'_m \right\}$
        \ENDIF
		\ENDFOR
  \STATE $I = {\rm Concat}({\rm Reshape}(\mathcal{T}))$ \ \ \ \   \% Calculating scores 
  \STATE ${\mathcal P} = \left\{ {{p_j}} \right\}_{j = 1}^N{\rm{ = NMS}}({\left\{ {{I_j}} \right\}_{j = 1}^N,\left\{ {{A_j}} \right\}_{j = 1}^N,\theta } )$
		\STATE \textbf{return} Part Set $\mathcal P$ 
	\end{algorithmic}  
\end{algorithm}

\subsection{Multi-Part and Multi-Scale Cross-Attention}
To effectively model the contextual dependencies between semantic parts at different scales and achieve mutual learning of global and local features, we design the MPMSCA module. First, we use the shared weight backbone as in Section \ref{sec:progressive} to extract part features and select the features of the last $A$ stages of $N$ discriminative parts. For the $n$-th part, the selected features are denoted as $\left\{f_{{p_n}}^{S-A+1}, f_{{p_n}}^{S-A+2}, \cdots, f_{{p_n}}^S \right\}$, where $1 \le n \le N$. As the features of different parts at the same stage are consistent in each dimension, thus for the $s$-th stage, we concatenate the features of different parts along the batch size direction for batch operations and obtain $f^s = \left[ {f_{{p_1}}^s;f_{{p_2}}^s;\cdots;f_{{p_N}}^s} \right]$, where $s\in \left\{S \! - \! A\! + \!1, S\! - \!A\! + \!2, \cdots,S\right\}$.

After that, $f^s$ are projected and flattened to $C^*$-dimensional tokens through the convolution block $B^s$ and global average pooling (GAP).  Followed by the rearrange operation, the number of parts $N$ is transferred from the batch size dimension to the sequence length dimension. The output $t^s \in {\mathbb {R}^{N \times C^*}}$ for the \textit{s}-th stage of all $N$ parts can be formulated as:
\begin{equation}
    t^s = {\mathop{\rm Rearrange}}\left( {\mathop{\rm GAP}\left( { B^{s}\left( {f^s} \right)} \right)} \right)
\end{equation}

Then the tokens representing different scales are concatenated to obtain the transformed tokens:
\begin{equation}
 {t_P} = \left\langle {t^{S-A +1}\parallel t^{S-A +2}\parallel \cdots \parallel t^S} \right\rangle 
\end{equation}
where $\left\langle {,\parallel ,} \right\rangle $ represents the concatenation operation along the sequence length direction, and ${t_P} \in {\mathbb {R}^{\left( {A \times N} \right) \times C^*}}$ is the fusion tokens containing multi-scale and multi-part information.

Interactive learning between detailed part descriptors and global representations is potentially useful in directing the network's attention toward subtle yet differentiated clues. To this end, we design a novel cross-attention mechanism operating at both object-part and part-part levels to improve FGVC performance. Specifically, 1×1 convolution is first applied to reduce the channel dimension of ${v_{im}^{S+1}}$, and we transform it to the same dimension as $t_P$, this process can be expressed as:
\begin{equation}
{t_{im}} = \varpi \left( { {{\rm{Con}}{{\rm{v}}_{1 \times 1}}\left( {v_{im}^{S+1}} \right)} } \right)
\end{equation}
where $\varpi$ represents the expand operation along the sequence length direction, and  ${t_{im}} \in {\mathbb {R}^{\left( {A \times N} \right) \times C^*}}$ denotes informative tokens containing global representations.

To obtain tokens containing the global and local information, we concatenate $t_{im}$ and $t_P$ along the channel direction and perform depthwise convolution. For the \textit{h}-th attention head, we multiply the obtained tokens and $t_P$ with the parameter matrix $W_h^Q, W_h^K,W_h^V$ to get the Queries $Q^h$, Keys $K^h$ and Values $V^h$, respectively:
\begin{gather}
    {Q^h} = {\rm DWConv}{ \left( {{{\rm Concat}}\left( {  {\varphi \left( {t_{im}} \right)} ,\varphi \left( {{\rm LN}({t_P})} \right)} \right)} \right)}W_h^Q \nonumber \\[1.5mm]
{K^h} = {t_P}W{_h^K}, \ {V^h} = {t_P}W_h^V
\end{gather}
where DWConv represents the depthwise convolution and $\varphi$ is linear projection function, LN denotes LayerNorm \cite{ba2016layer}.

% The reorganized $Q^h$ and $K^h$ are used to calculate the attention map $A^h$ of a single attention head.
To suppress the noise and reduce the impact of redundant information, we define a mask ${\mathcal{M}^h}$ based on the attention map $A^h$. This process is expressed as:
\begin{equation}
    {A^h} = {\rm{Softmax}}\left( {\frac{{{Q^h}{(K^h)}^\top}}{{{\left( {{C^*}/H} \right)^{1/2}} }} \odot {\mathcal{M}^h}} \right) = \left[ {a_1^h,a_2^h,\cdots,a_L^h} \right]
\end{equation}
with the binary mask matrix ${\mathcal{M} ^h}$ is defined as:
\begin{equation}
{\mathcal{M}}_{\left( {i,j} \right)}^h = \left\{ {\begin{array}{*{20}{l}}
  {1}&{{\rm{if}}\;A_{(i,j)}^h\;{\rm{is}}\; {\rm{top}}\!-\! v\;{\rm{value}}\;,}\\
{0}&{{\rm{otherwise}}.}
\end{array}} \right.
\end{equation}
where $h \in$ $\left\{{1,2,...,H}\right\}$ and $H$ is a hyper-parameter representing the number of heads, ${\left( {{C^*}/H} \right)^{1/2}}$ is a scaling factor, $\odot$ denotes element-wise dot product, $a^h_j \in \mathbb {R}^L$ represents the similarities between the \textit{j}-th token and the others in the \textit{h}-th head, and $v$ is a hyper-parameter denoting the number of selected tokens in the $i$-th row.

The outputs of single-head attentions are concatenated and linearly projected to get the final output:
\begin{equation}
\begin{aligned}
    {U^h}& = {A^h} \otimes {V^h} \\[1.5mm]
Y = \left( {1 - \beta } \right) {\rm Concat}  & \left( {{U^1},{U^2},\cdots,{U^H}} \right){W^O}  + \beta {t_P}
\end{aligned}
\end{equation}
where $\otimes$ represents matrix multiplication, $U^h$ is the output of \textit{h}-th attention head, $W^O$ denotes the weight of the linear projection layer, $\beta$ is a learnable parameter that adaptively adjusts the weight ratio and $Y\in {\mathbb {R}^{\left( {A \times N} \right) \times C^*}}$ denotes a powerful feature that fully considers the long-range correlation between multiple parts and multiple scales.

Noted that, the cross-attention mechanism preserves the order of the input sequence, and it's observed that the classification prediction ability of deep layers tends to outperform that of shallow layers. It is not inappropriate to treat deep and shallow layers equally for predictions. Thus we rearrange and split the output $Y$ to regain part tokens representing different scales $\hat t_P^s \in {\mathbb {R}^{N \times C^*}}$, and add them to $t^s$ through residual connection, which can be described as:
\begin{gather}
    \hat t_P^{S- A+1},t_P^{S- A+2},\cdots,\hat t_P^S = {\rm{Avg}}\left( {{\rm{Split}}\left( {{\rm{Rearrange}}\left( Y \right)} \right)} \right) \nonumber \\[1.5mm]
    v_P^s = {\rm{ Rearrange}}\left( {\hat t_P^s + t^s} \right)
\end{gather}
where Rearrange$(\cdot )$, Split$(\cdot )$, and Avg$(\cdot )$ denote the operations of dimension rearrangement, token split and averaging, respectively, and $v_P^s \in {\mathbb {R}^ {C^*}}$ represents the feature vector of all parts in the $s$ stage, where $s \in \left\{S \! - \! A\! + \!1, S\! - \!A\! + \!2, \cdots, S\right\}$. 

Like ${v_{im}^{S+1}}$, we concatenate the part feature vectors at the last $A$ stages $v_{P}^{S - A+1}$, $v_{P}^{S - A+2}$, $\cdots$, $v_{P}^{S}$ to obtain the multi-scale fusion vector ${v_{P}^{S+1}}$, then feed the part feature vectors of the last $A$ stages and multi-scale fusion vector to the corresponding shared weight QP classifier to get the predictions $ \hat y_{P1}^{s'}$ and $\hat y_{P2}^{s'}$, where $s'\in \left\{S \! - \! A\! + \!1, S\! - \!A\! + \!2, \cdots, S+1\right\}$. The loss functions calculated in Section \ref{sec:progressive} are repeated for the part vectors. Similar to Eq. (\ref{equ:loss}), the total classification loss for all parts at all scales can be calculated as:
\begin{equation}
 {{\mathcal L}_{part}} = \sum\limits_{s' = S-A+1}^{S + 1} {\left[ {{{\ell}_{sce}}(\hat y_{P1}^{s'},{y_{{\alpha ^{s'}}}}) + {\ell _{ce}}(\hat y_{P2}^{s'},y) + \ell _{re{g^{P}}}^{s'}} \right]} 
\end{equation}

\subsection{Training and Inference}
Based on the above-proposed modules, our framework is easy to implement and can be trained in an end-to-end manner. To be specific, the overall training loss is below: 
\begin{equation}
    \mathcal{L}=\mathcal{L}_{im}+\mathcal{L}_{part}
\end{equation}

In the test phase, we remove the auxiliary classifiers in all quality probing classifiers as well as the entire part branch (i.e., part navigator and MPMSCA module), and only use the image branch for prediction. To consistently enhance the model's generalization ability, we combine the predictions from all main classifiers to generate the final prediction:
\begin{equation}
    \hat y_{im}^{final} = \sum\limits_{s' = S-A+1}^{S + 1} \hat y_{im1}^{s'}
\end{equation}

To sum up, the computational cost of our CSQA-Net during inference is slightly higher than that of baseline networks (e.g., ResNet50 or Swin-Base).

\section{EXPERIMENTS}
\label{sec:experiment}
In this section, we first introduce four FGVC benchmark datasets in Section \ref{sec:Datasets}, and we analyze our experimental settings in Section \ref{sec:details}, including various configurations and hyper-parameters. After that, we compare the proposed method with the current state-of-the-art methods, and the results are illustrated in Section \ref{sec:compare}. To further explain the impact of each component and the effect of hyper-parameters, performance analysis is conducted in Section \ref{sec:perform}. Last but not least, in Section \ref{sec:visual}, the visualization results help to understand the motivation of our approach.

\begin{table}[!t]
\renewcommand{\arraystretch}{1.2}
\caption{Statistics of four benchmark datasets for fine-grained visual categorization.\label{tab:datasets}}
\centering
\begin{tabular}{|c |c |c |c|}
\hline
Dataset & Category & Training &  Testing \\
\hline
\hline
CUB-200-2011 \cite{CUB} & 200 & 5994 & 5794 \\
Stanford Cars \cite{CAR} & 196 & 8144 & 8041 \\
FGVC-Aircraft \cite{maji2013fine} & 100 & 6667 & 3333 \\
NABirds \cite{NAB} & 555 & 23,929 &24,633 \\
\hline
\end{tabular}
\end{table}

\subsection{Datasets}
\label{sec:Datasets}
To verify the robustness of our proposed CSQA-Net, we conduct comprehensive experiments on four public benchmarks for fine-grained recognition, including CUB-200-2011 (CUB) \cite{CUB}, Stanford Cars (CAR) \cite{CAR}, FGVC-Aircraft (AIR) \cite{maji2013fine} and NABirds (NAB) \cite{NAB}. As shown in Tab. \ref{tab:datasets}, we use a standard method to split training/testing images, with top-1 accuracy serving as the evaluation metric. Throughout the experiments, we solely utilize image labels for supervised training, without employing any additional bounding boxes or part annotations. 

\begin{table}[ht!]
\renewcommand{\arraystretch}{1.2}
  \begin{center}
    \caption{Comparison with different methods on CUB-200-2011. \label{tab:cub}}
    % \resizebox{\linewidth}{!}{
    \begin{tabular}{|c|c|c|c|} % <-- Alignments: 1st column left, 2nd middle and 3rd right, with vertical lines in between
      \hline
       Method& Publication & Backbone & Acc.@1 (\%)\\
       \hline
       \hline
       S3Ns \cite{ding2019selective}& ICCV 2019& ResNet-50& 88.5\\
       ACNet \cite{ji2020attention} & CVPR 2020 & ResNet-50& 88.1 \\
 API-Net \cite{zhuang2020learning}& AAAI 2020 & DenseNet-161&90.0\\
 PMG \cite{PMG}& ECCV 2020& ResNet-50&89.6\\
       AKEN \cite{9023386}& TCSVT 2021 & VGG-19 & 87.1 \\
       AP-CNN \cite{ding2021ap}& TIP 2021 & ResNet-50& 88.4 \\
       DP-Net \cite{wang2021dynamic}& AAAI 2021& ResNet-50& 89.3\\
 PART \cite{9614988}& TIP 2021& ResNet-50&89.6\\
       DPN \cite{9389760}& TCSVT 2022 & ResNet-50& 89.1 \\
 P-CNN \cite{P-CNN}& TPAMI 2022& VGG-19&87.3\\
 CMN \cite{deng2022fine}& TIP2022& ResNet-50&88.2\\
       P2P-Net \cite{yang2022fine}& CVPR 2022 & ResNet-50& 90.2 \\
 ME-ASN \cite{ME-ASN}& TMM 2022& ResNet-50&89.5\\
       GDSMP-Net \cite{ke2023granularity}& PR 2023 & ResNet-50& 89.9 \\
 TA-CFN \cite{TA-CFN}& TNNLS 2023& ResNet-50&\textbf{90.5}\\
       \rowcolor{black!10} CSQA-Net (Ours)& n/a& ResNet-50&\textbf{90.5}\\
 \hline
        TransFG \cite{he2022transfg}& AAAI 2022&  ViT-B-16& 91.7\\
 ViT-NeT \cite{Vit-net}& ICML 2022& Swin-B&91.6\\
 DCAL \cite{DCAL}& CVPR 2022& R50-ViT-B&92.0\\
 IELT \cite{10042971}& TMM 2023 & ViT-B-16 &91.8\\
        AA-trans \cite{AA-trans}&  PR 2023& ViT-B-16&91.4\\
 TransIFC+ \cite{Transifc}& TMM 2023& Swin-B&91.0\\
        Dual-TR \cite{Dual-TR}& TCSVT 2023& ViT-B-16& 92.0\\
 ACC-ViT \cite{ACC-ViT}& PR 2024& ViT-B-16&91.8\\
       \rowcolor{black!10} CSQA-Net (Ours)& n/a & Swin-B& \textbf{92.6}\\
        \hline
    \end{tabular}
    % }
  \end{center}
  % \vspace{-10pt}
\end{table}
\subsection{Implementation Details}
\label{sec:details}
We use ResNet-50 \cite{he2016deep} and Swin Transformer (Swin-Base) \cite{liu2021swin} pre-trained on ImageNet \cite{deng2009imagenet} as the backbone network. As commonly used augmentations \cite{PMG,wang2023accurate} in FGVC, random cropping and horizontal flipping are applied for training, while center cropping is used for testing, and finally all input images are resized to 448×448 resolution. We also utilize a stochastic gradient descent (SGD) optimizer with a momentum of 0.9 and weight decay 5e-4 to optimize the training process. The initial learning rate is 0.0075 for the CUB, AIR and NAB datasets, and 0.01 for the CAR dataset, and scheduled by cosine annealing strategy \cite{loshchilov2016sgdr}. Moreover, the learning rate of the auxiliary classifier is set constant to 0.01. Notably, our model is computationally efficient and only needs to be trained for 50 epochs (including 5 warm-up epochs), with a batch size of 16. Experiments are conducted using PyTorch version 1.7 or higher on Nvidia GeForce RTX 3090 GPUs, and our model is trained end-to-end in a unified scheme.

\begin{table}[ht!]
\renewcommand{\arraystretch}{1.2}
  \begin{center}
    \caption{Comparison with different methods on Stanford Cars. \label{tab:car}}
    % \resizebox{\linewidth}{!}{
    \begin{tabular}{|c|c|c|c|} % <-- Alignments: 1st column left, 2nd middle and 3rd right, with vertical lines in between
      \hline
       Method& Publication & Backbone & Acc.@1 (\%)\\
       \hline
       \hline
       DCL \cite{chen2019destruction}& CVPR 2019& ResNet-50& 94.5\\
 API-Net \cite{zhuang2020learning}& AAAI 2020 & DenseNet-161&95.3\\
 SPS \cite{huang2021stochastic}& ICCV 2021& ResNet-50&94.4\\
 AKEN \cite{9023386}& TCSVT 2021 & VGG-19 &93.9\\
 P2P-Net \cite{yang2022fine}& CVPR 2022 & ResNet-50&95.4\\
 P-CNN \cite{P-CNN}& TPAMI 2022& VGG-19&93.3\\
 CMN \cite{deng2022fine}& TIP 2022& ResNet-50&94.9\\
       DPN \cite{9389760}& TCSVT 2022 & ResNet-50& 95.0\\
 PMRC \cite{tang2023weakly}& CVPR 2023& ResNet-50&95.4\\
 SIA-Net \cite{wang2023semantic}& TCSVT 2023 & ResNet-50&95.5\\
        \rowcolor{black!10} CSQA-Net (Ours)& n/a& ResNet-50&\textbf{95.6}\\
 \hline
        SwinTrans \cite{liu2021swin}& ICCV 2021&  Swin-B & 94.2\\
        DCAL \cite{DCAL}& CVPR 2022& R50-ViT-B&95.3\\
        TransFG \cite{he2022transfg}& AAAI 2022& ViT-B-16&94.8\\
 ViT-NeT \cite{Vit-net}& ICML 2022& Swin-B&95.0\\
 MpT-Trans \cite{MpT-Trans}& ACM MM 2023& ViT-B-16&93.8\\
 ACC-ViT \cite{ACC-ViT}& PR 2024& ViT-B-16&94.9\\
        \rowcolor{black!10} CSQA-Net (Ours)& n/a & Swin-B& \textbf{95.6}\\
        \hline
    \end{tabular}
  % }
  \end{center}
\end{table}
For hyper-parameter fine-tuning, we set the number of parts $N$ and the last $A$ stages of the backbone to 4 and 3 respectively. The channel number for the convolutional projection in the MLSQE module is $C^*=$ 1024. We empirically set the smoothing factors $\alpha^{s'}$ to $\left\{0.7, 0.8, 0.9, 1.0 \right\}$ in ascending order, as intuitively deeper layers of the network are expected to make more confident predictions. Both $I$ and $\lambda$ in the quality probing classifier are set as 2, and $\varepsilon^{s'}$ is defined as half of  $\alpha^{s'}$. Additionally, the number of attention heads $H$ in the MPMSCA module and the $v$ in the mask $\mathcal{M}$ are set to 16 and 3 respectively. These optimal hyper-parameter settings will be further discussed in Section \ref{sec:perform}.

\subsection{Comparison With the \textit{State-of-the-Arts }}
\label{sec:compare}

\subsubsection{CUB-200-2011}
As shown in Tab. \ref{tab:cub}, we divide the methods into two groups from top to bottom: weakly supervised CNN-based networks and Transformer-based methods. It can be observed that CNN-based approaches such as API-Net \cite{zhuang2020learning} and P2P-Net \cite{yang2022fine} consistently enhance performance to over 90{\%} without adding annotations. However, they ignore the feature interaction between discriminative parts and the real-time evaluation of feature quality. Considering these two aspects, our method achieves a new state-of-the-art result of 90.5{\%} on the ResNet50 backbone. Notably, TA-CFN \cite{TA-CFN} achieves comparable performance to ours, but it relies on additional text-assisted information. In contrast, we only use image-level labels to supervise model training. Furthermore, our proposed MPMSCA module exhibits a certain correlation with the self-attention mechanism, which is the core component of the Transformer. Consequently, we also verify the robustness of our model under the Swin-Base backbone. Note that, our proposed CSQA-Net outperforms existing methods such as Dual-TR \cite{Dual-TR} and ViT-NeT \cite{Vit-net}, achieving a high performance of 92.6{\%}. We attribute these significant improvements to the proposed key modules, which guide the network to discover discriminative clues in objects.
\begin{table}[ht!]
\renewcommand{\arraystretch}{1.2}
  \begin{center}
    \caption{Comparison with different methods on FGVC-Aircraft.\label{tab:air}}
    % \resizebox{\linewidth}{!}{
    \begin{tabular}{|c|c|c|c|} % <-- Alignments: 1st column left, 2nd middle and 3rd right, with vertical lines in between
      \hline
       Method& Publication & Backbone & Acc.@1 (\%)\\
       \hline
       \hline
       PMG \cite{PMG}& ECCV 2020 & ResNet-50& 93.4 \\
 API-Net \cite{zhuang2020learning}& AAAI 2020 & DenseNet-161&93.9\\
       CAL \cite{CAL}& ICCV 2021& ResNet-50&94.2\\
       DP-Net \cite{wang2021dynamic}& AAAI 2021& ResNet-50& 93.9 \\
       DPN \cite{9389760}& TCSVT 2022 & ResNet-50& 93.2 \\
 CMN \cite{deng2022fine}& TIP2022& ResNet-50&93.8\\
       P2P-Net \cite{yang2022fine}& CVPR 2022 & ResNet-50& 94.2 \\
 GDSMP-Net \cite{ke2023granularity}& PR 2023 & ResNet-50&94.4\\
       SIA-Net \cite{wang2023semantic}& TCSVT 2023 & ResNet-50& 94.3 \\
 \rowcolor{black!10} CSQA-Net (Ours)& n/a& ResNet-50&\textbf{94.5} \\
       \hline
        SwinTrans \cite{liu2021swin}& ICCV 2021& Swin-B&92.2\\
 TransFG \cite{he2022transfg}& AAAI 2022& ViT-B-16&94.0\\
 DCAL \cite{DCAL}& CVPR 2022& R50-ViT-B&93.3\\
 MpT-Trans \cite{MpT-Trans}& ACM MM 2023& ViT-B-16&92.2\\
 ACC-ViT \cite{ACC-ViT}& PR 2024& ViT-B-16&93.5\\
        \rowcolor{black!10} CSQA-Net (Ours)& n/a & Swin-B& \textbf{94.7}\\
        \hline
    \end{tabular}
  % }
  \end{center}
\end{table}

\subsubsection{Stanford Cars}
The comparison results of different methods on the Stanford Cars dataset are presented in Tab. \ref{tab:car}. Despite the minor fluctuations in recognition accuracy among various methods on this dataset, our proposed CSQA-Net still surpasses existing highly competitive methods on the ResNet50 backbone. Notably, compared with PMRC \cite{tang2023weakly}, which requires 600 epochs for training, our method is easy to implement and achieves a SOTA performance of 95.6{\%}. When considering Transformer-based methods, DCAL\cite{DCAL} designs cross-attention between global key values and local queries, which achieves 95.3{\%} top-1 accuracy. Different from DCAL, we exploit feature descriptors of multi-scale parts and global objects to explore richer details and texture information and thus exceed it by 0.3{\%} top-1 accuracy.

\subsubsection{FGVC-Aircraft}
In Tab. \ref{tab:air}, we compare existing high-performance methods with our network on the FGVC-Aircraft dataset. As can be seen, our proposed CSQA-Net outperforms existing top methods and achieves a new result of 94.5{\%} on the ResNet50 backbone, whose performance improvement can be attributed to our proposed components. Furthermore, by substituting the ResNet50 backbone with the more advanced Swin-Base, our CSQA-Net achieves 94.7{\%} top-1 accuracy and exhibits excellent recognition results, which is 1.4{\%} and 1.2{\%} higher than recent DCAL \cite{DCAL} and ACC-ViT \cite{ACC-ViT} respectively, validating the effectiveness of our designed approach.

\subsubsection{NABirds}
The recognition performance on the NABirds dataset is listed in Tab. \ref{tab:nab}. It is evident that recent CNN-based methods, such as GDSMP-Net \cite{ke2023granularity} and GHORD \cite{zhao2021graph}, achieve performance of 89.0{\%} and 88.0{\%}, respectively, easily surpassing previous work such as MaxEnt \cite{maxent}. For a fair comparison, we also adopt the ResNet50 backbone and achieve the highest performance of 90.0{\%}. To be noticed, Transformer-based methods demonstrate more competitive results on larger datasets. With the more robust Swin-Base backbone, our proposed CSQA-Net further improves accuracy to a new level of 92.3{\%}. These results validate that the modules we designed can be well integrated with existing hierarchical Transformer architecture and steadily improve performance.

\begin{table}[!t]
\renewcommand{\arraystretch}{1.2}
\caption{Comparison with different methods on NABirds. \label{tab:nab}}
\centering
\begin{tabular}{|c |c |c |c|}
\hline
Method& Publication & Backbone & Acc.@1 (\%)\\
\hline
\hline
MaxEnt \cite{maxent}&  NIPS 2018 & DenseNet-161 & 83.0 \\
DSTL \cite{cui2018large}& CVPR 2018 & Inception-v3 & 87.9 \\
Cross-X \cite{luo2019cross}& ICCV 2019 & ResNet-50 & 86.4 \\
API-Net \cite{zhuang2020learning}& AAAI 2020 & DenseNet161 & 88.1 \\
GHORD \cite{zhao2021graph}& CVPR 2021 & ResNet-50 & 88.0 \\
 SR-GNN \cite{SR-GNN}& TIP 2022& ResNet-50&88.8\\
 GDSMP-Net \cite{ke2023granularity}& PR 2023 & ResNet-50 &89.0 \\
 LGTF \cite{LGTF}& ICCV 2023& ResNet-50&89.5\\
\rowcolor{black!10}CSQA-Net (Ours)& n/a & ResNet-50 &\textbf{90.0}\\
\hline
TransFG \cite{he2022transfg}& AAAI 2022 & ViT-B-16 & 90.8 \\
IELT \cite{10042971}& TMM 2023 & ViT-B-16 & 90.8 \\
 TransIFC+ \cite{Transifc}& TMM 2023& Swin-B&90.9\\
 Dual-TR \cite{Dual-TR}& TCSVT 2023& ViT-B-16&91.3\\
 ACC-ViT \cite{ACC-ViT}& PR 2024& ViT-B-16&91.4\\
 FET-FGVC \cite{FET-FGVC}& PR 2024& Swin-B&91.7\\
\rowcolor{black!10}CSQA-Net (Ours)& n/a & Swin-B & \textbf{92.3}\\
\hline
\end{tabular}
\end{table}

\begin{table}[!t]
\renewcommand{\arraystretch}{1.2}
\caption{Ablation study on the proposed components with the baseline ResNet-50. Part Nav: part navigator. MLSQE: multi-level semantic quality evaluation. MPMSCA: multi-part and multi-scale cross-attention.
\label{tab:ablation}}
\centering
\begin{tabular}{|c |c |c |c |c |c |c|}
\hline
\multirow{2}*{PART NAV}&  \multirow{2}*{MLSQE} & \multirow{2}*{MPMSCA}& \multicolumn{4}{c|}{Acc.@1 (\%)} \\ 
\cline{4-7}
         &  &  & CUB & CAR & AIR & NAB \\
\hline
\hline
- & - & - & 85.4&92.7& 90.3& 84.6\\
\checkmark & - & - & 86.5&93.1& 91.2& 85.7\\
 -& \checkmark& -& 88.5& 94.3& 93.0&88.0\\
\checkmark & \checkmark & - & 89.2&94.8& 93.6& 88.8\\
\checkmark & \checkmark & \checkmark & 90.5&95.6& 94.5& 90.0\\
\hline
\end{tabular}
\end{table}

\subsection{Performance Analysis}
\label{sec:perform}

\subsubsection{Ablation Studies on Different Components}
To demonstrate the robustness of our proposed components, we conduct ablation studies on four widely used FGVC benchmarks. As shown in Tab. \ref{tab:ablation}, the first row lists the classification accuracy of the pure baseline ResNet50, employing the same experimental configuration and training parameters as our final model. First, when we only add the Part Navigator to mine the discriminative parts and use the same backbone and classifier as the baseline for prediction, the performance improves by roughly 1{\%} across all four datasets. After further adding the MLSQE module, the network effectively leverages multi-level complementary semantics and encourages them to be more discriminative for boosting classification accuracy, e.g., from 86.5{\%} to 89.2{\%} on the CUB dataset, and from 91.2{\%} to 93.6{\%} on the AIR dataset. Comparing the results of the last two rows, the designed MPMSCA module elevates the classification results significantly, as it considers more detailed part descriptors and their interaction with global information, prompting the network to mine subtle but distinctive details. Notably, the third row indicates that we have discarded the part navigator and MPMSCA module, which will be further discussed in Section \ref{sec:n}. In short, the above results verify the robustness of the crucial components in CSQA-Net.

\subsubsection{Effect of Quality Probing (QP) Classifier}
In Tab. \ref{tab:hyper}, we show the influence of different hyper-parameter configurations on the QP classifier. Our proposed classifier is primarily determined by two hyper-parameters, which are the epoch initialization parameter $\delta$ and the intensity adjustment factor $\lambda$. It can be found that as the value of $\delta$ increases, the performance initially increases and then decreases. This is because a smaller $\delta$ hampers the auxiliary classifier's ability to fit the data well, whereas a larger $\delta$ will hinder the auxiliary classifier from effectively reflecting the quality of the sample. Furthermore, $\lambda$ governs the strength adjustment of the regularization term. Although the disparities shown in the table are marginal, appropriate $\lambda$ can also lead to performance enhancements for FGVC. Experimental findings suggest that setting $\delta$ = 2 and $\lambda$ = 2  consistently yields robust classification performance across datasets. Moreover, when we substitute the QP classifier with a 2-layer MLP constrained by cross-entropy loss, the accuracy decreases from 90.5{\%} to 88.9{\%} on the CUB dataset, which verifies the significance of our proposed QP classifier for regularization representation.

\begin{table}[!t]
\renewcommand{\arraystretch}{1.2}
\caption{Hyper-parameter experiments on quality probing classifier on CUB-200-2011. Acc.@1 (\%). \label{tab:hyper}}
\centering
\begin{tabular}{|c | c c c c c|}
\hline
Value & $\delta$=1 & $\delta$=2 & $\delta$=3 & $\delta$=4  &$\delta$=5\\
\hline
\hline
$\lambda$=1& 89.7& 90.2& 89.8& 89.7&89.5\\

$\lambda$=2& 89.8& \textbf{90.5}& 90.0& 89.8&89.7\\

$\lambda$=3& 89.5& 90.0& 90.1& 89.6&89.4\\

$\lambda$=4& 89.6& 89.8& 89.6& 89.3&89.3\\
\hline
\end{tabular}
\end{table}

\begin{figure}[!t]
\centering
\includegraphics[width=1\linewidth]{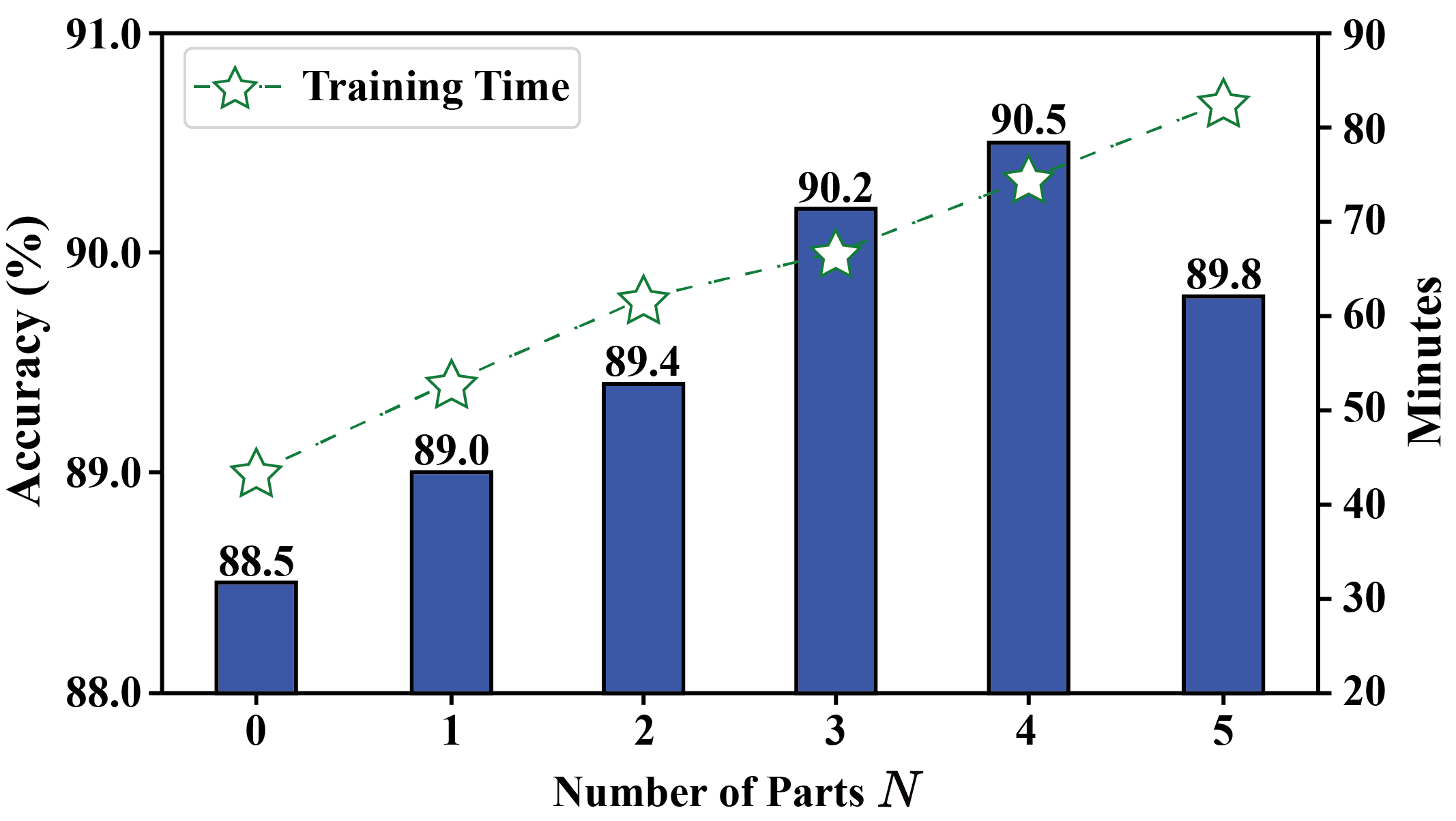}
\caption{Influence of the Number of Parts $N$ on CUB-200-2011.}
\label{fig: number of parts}
\end{figure}

\subsubsection{Influence of The Number of Parts}
\label{sec:n}
As depicted in Fig. \ref{fig: number of parts}, with the increases in the number of parts $N$, the overall trend of classification accuracy exhibits an initial ascent followed by a decline. When $N$ = 4, the model achieves its peak accuracy of 90.5{\%} on the CUB dataset. Although smaller values of $N$ can reduce training and inference time, they also result in fewer captured discriminative details, thereby impeding the provision of diverse prior knowledge for the MPMSCA module. However, an excessively large value of $N$ may introduce additional redundancy and noise, leading to diminished performance. To be noticed, we also test the case of $N$ = 0, which means that we completely remove the part branch (i.e., part navigator and MPMSCA module) and solely utilize the image branch for training. This corresponds to the results displayed in the third row of Tab. \ref{tab:ablation}.

\subsubsection{Influence of Hyper-Parameters on MPMSCA Module}
As Tab. \ref{tab:MPMSCA} demonstrates, we adopt control variates to assess the impact of different hyper-parameters on multi-part and multi-scale cross-attention module. Specifically, increasing the value of $C^*$ fails to improve performance as the additional convolutional projection disrupts the original relationship between image channels. Moreover, a proper value of $H$ can enhance feature representation capabilities, facilitating the model's generalization to unknown data. In addition, as $v$ increases, the accuracy increases first and then decreases. If $v$ is too small, the potential of the MPMSCA module is not fully exploited to model dependencies between tokens. Conversely, an excessively large value for $v$ deviates from the original intention of introducing the mask matrix $M$. Note that, the MPMSCA module uses the same hyper-parameters across different backbone networks (i.e., ResNet50 and Swin-Base) for a fair comparison, and this module is only activated during the training phase.

\begin{table}[!t]
\renewcommand{\arraystretch}{1.3}
\caption{Hyper-parameter experiments on MPMSCA module on CUB-200-2011. $C^*$ and $H$ represent the projected channel dimension and the head number of cross-attention respectively. $v$ denotes the top-$v$ value in the mask $M$.\label{tab:MPMSCA}}
\centering
\begin{tabular}{|c |c c c c c c|}
\hline
Channel ($C^*$)& 1024  & 1024& 2048& 1024& 1024 &1024\\
Head ($H$)    & 8     & 8    & 8& 16    & 16 &32\\
top-$v$ ($v$) & 2     & 3& 3     & 3    & 4 &4\\
\hline
\hline
Acc.@1 (\%)& 89.9& 90.2& 89.8& \textbf{90.5}& 90.1 &90.0\\
\hline
\end{tabular}
\end{table}

\begin{table}[!t]
\renewcommand{\arraystretch}{1.3}
\caption{Ablation study of different variants of MPMSCA module on CUB-200-2011. \label{tab:variants}}
\centering
\begin{tabular}{|c |c c c c|}
\hline
Variants & w/o MPMSCA& w SA& w CA&w MPMSCA\\
\hline
\hline
Acc.@1 (\%)&89.2& 89.7& 90.1&\textbf{90.5}\\
\hline
\end{tabular}
\end{table}

\subsubsection{Effect of Different Variants of The MPMSCA Module}
To further validate the efficacy of the MPMSCA module, we conduct experiments using various alternative methods in Tab. \ref{tab:variants}. Here, SA and CA represent instances where the MPMSCA module is replaced by Self-Attention and pure Cross-Attention mechanism, respectively. The global image or local part tokens they utilize are derived from the deepest output of the backbone network. While some improvements were observed, relying solely on self-attention operations between detected salient parts proved insufficient, as it neglects global semantic information. By introducing cross-attention between global image queries and multi-part key values, performance was further elevated to 90.1\%. However, the rich information embedded in multi-scale features remained underutilized. Through the proposed MPMSCA module, global object and multi-scale part information are amalgamated to form a novel query vector, subsequently employed to conduct cross-attention with key values composed of part tokens, thereby distributing attention to discriminative details. Therefore, the recognition performance achieves a new level of 90.5{\%}.

\begin{figure*}
    \centering
    \includegraphics[width=1\linewidth]{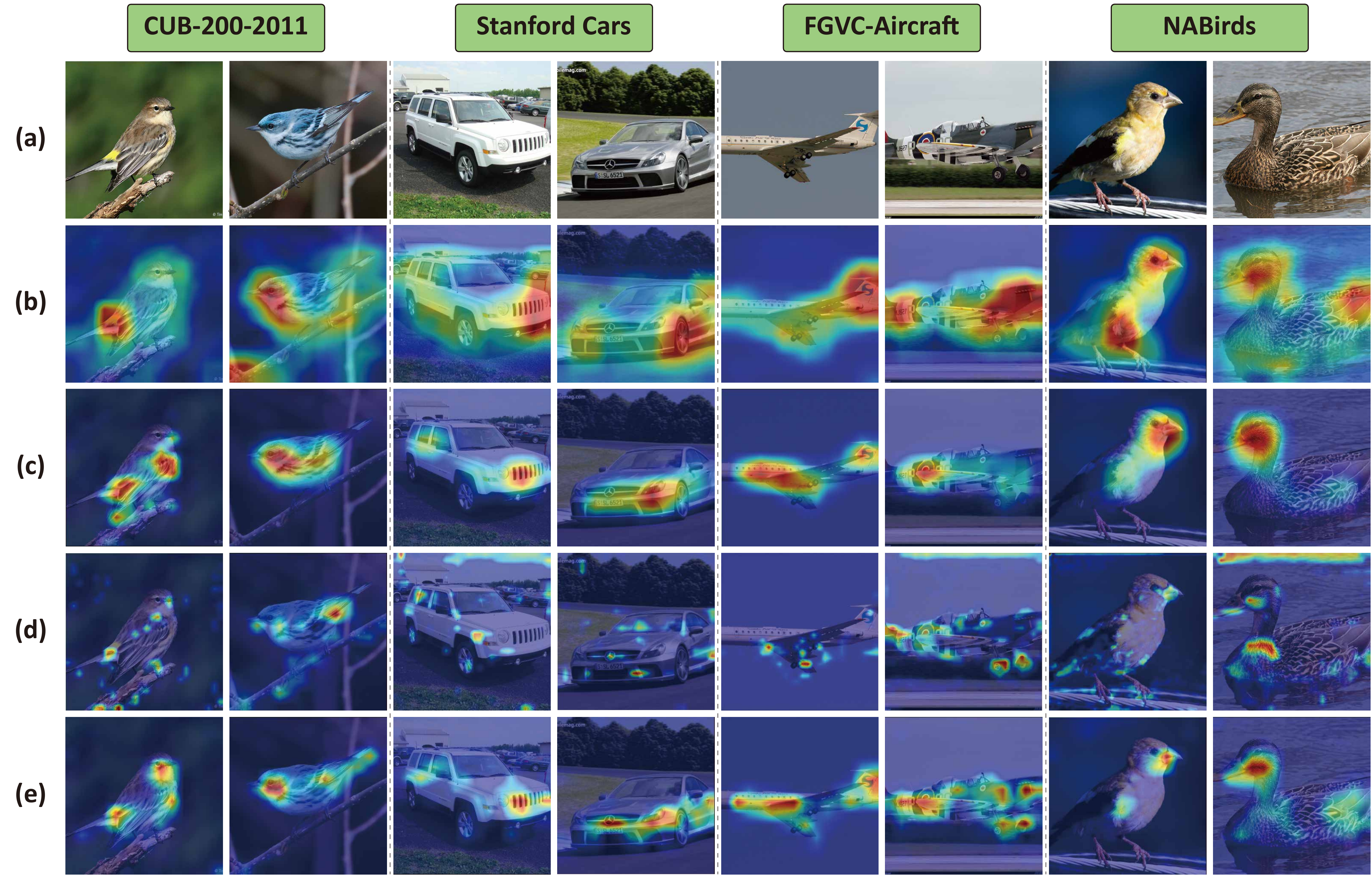}
    \caption{Visualization results of our CSQA-Net on different datasets. (a) shows the input images. (b) and (c) represent the heatmaps generated by baseline and baseline+MLSQE, respectively. (d) and (e) are the visualization results of feature maps at Stage S-1 and S of our method, respectively.}
    \label{fig:visual}
\end{figure*}

\subsubsection{Effect of selecting the last \texorpdfstring{$A$}{1} stages of backbone}
As listed in Tab. \ref{tab:stage}, selecting a smaller number of stages cannot discover and mine low-level detailed information such as texture and edges, and the MLSQE and MPMSCA modules are not fully utilized to improve fine-grained recognition. However, it is not always advantageous to select too many stages. The feature maps output by stages close to the input image (i.e., the first and second stages) have relatively small spatial receptive fields and lack sufficient semantic information, thus they can only express simple image structures and fail to explore discriminative details in fine-grained images. To achieve a trade-off between memory consumption and efficient computation, we set $A$ to 3 as our final experimental setting.

\begin{table}[!t]
\renewcommand{\arraystretch}{1.3}
\caption{Ablation study on the effect of $A$ on CUB-200-2011. \label{tab:stage}}
\centering
\begin{tabular}{|c |c c c c|}
\hline
Values of $A$& 1& 2& 3     & 4\\
\hline
\hline
Acc.@1 (\%)& 88.2& 89.8& \textbf{90.5}& 90.2\\
\hline
\end{tabular}
\end{table}
 
\subsection{Visualization Analysis}
\label{sec:visual}

\subsubsection{Class Activation Map}
Fig. \ref{fig:visual} presents the intuitive visualization results of our CSQA-Net method on four benchmark datasets. Row (a) showcases two randomly sampled images from each dataset. Comparing the baseline model in row (b) and the model in row (c) that only adds the MLSQE module, it can be found that baseline tends to activate both the entire object and background information that interferes with classification. Conversely, the model enhanced with the MLSQE module accurately identifies class-discriminative regions of the object. Rows (d) and (e) show the activation maps of the feature maps output by the CSQA-Net at the penultimate and final stages of the backbone. Note that, the deep semantic information highlights subtle yet differentiated regions between objects, with minimal noise background activation. This also corresponds to the fact that we should give higher confidence (a larger $\alpha$) to high-level features. Moreover, to validate the effectiveness of the part navigator and MPMSCA module, we can individually compare rows (c) and (e). With the addition of the part branch, the model can discern discarded details and focus on small-scale components rather than entire connected regions seen in row (c), e.g., the head and wings of a bird, the lights and doors of a car, the fuselage, wings and wheels of an aircraft. These visualization results fully verify and demonstrate the robustness of each proposed component. 

\begin{figure}[!t]
\centering
\includegraphics[width=1\linewidth]{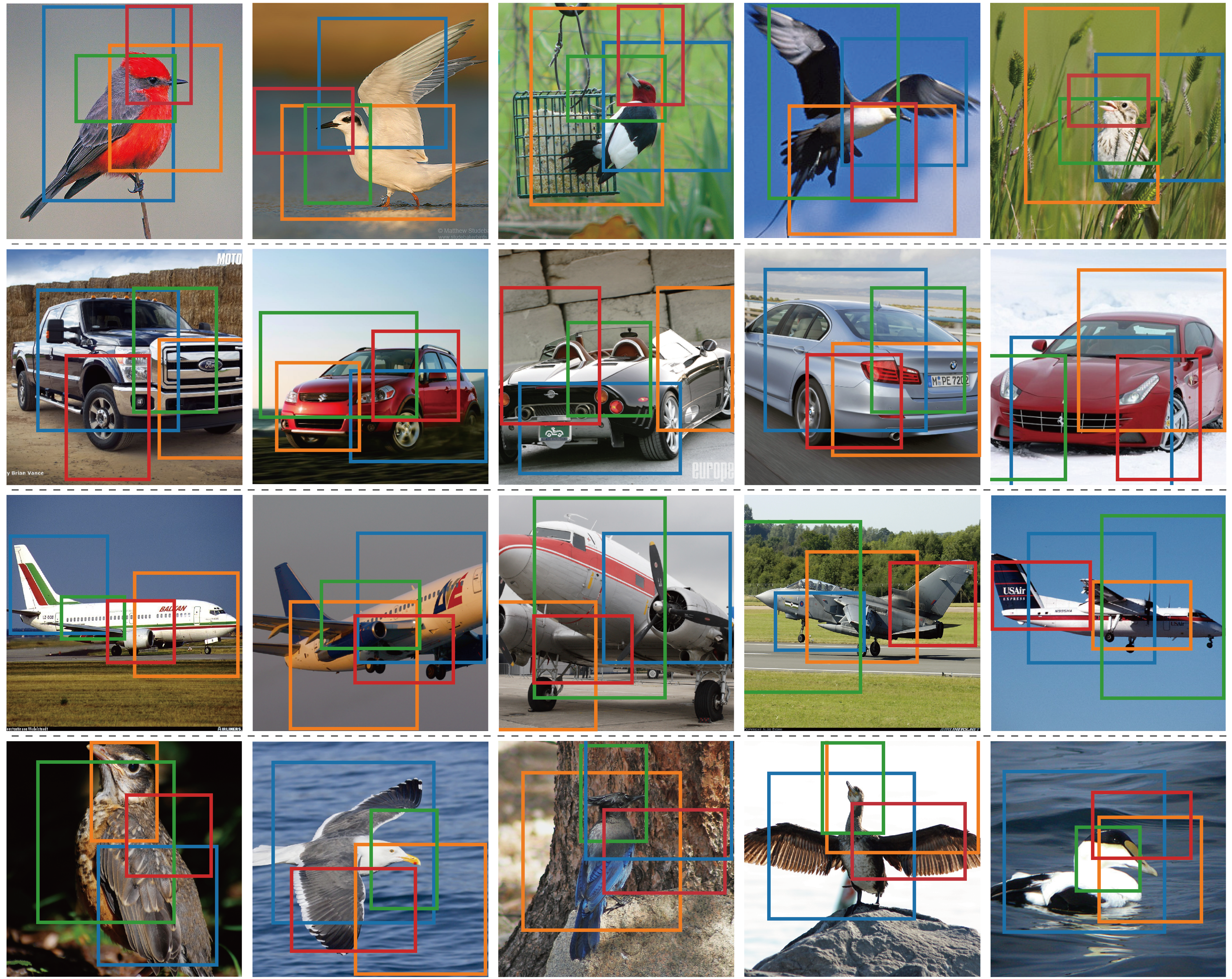}
\caption{Visualization of discriminative regions generated by the Part Navigator on different fine-grained images. }
\label{fig: discriminative regions}
\end{figure}

\subsubsection{Located Salient Regions}
To verify the part location ability of the part navigator, we visualize top-4 scoring bounding boxes generated by it in Fig. \ref{fig: discriminative regions}. The first row shows various species of birds from the CUB dataset. Despite the presence of noise background, these part boxes accurately capture the discriminative regions of birds with different postures, i.e., the head, body and wings of birds. For rigid objects in the CAR and AIR datasets, the regions detected by the part navigator module align well with our intuitive perception, that is, the headlamps and body of cars, and the fuselage and wings of aircraft. Moreover, the results on the last row of the NAB dataset are similar to CUB since both are fine-grained benchmarks about birds. Overall, these detection results validate the efficacy of the proposed part navigator and furnish crucial region priors knowledge for the MPMSCA module.

\section{Conclusion}
\label{sec:conclude}
In this paper, we propose a novel FGVC architecture, named CSQA-Net, which utilizes discriminative part features to regularize global semantics across multiple granularities and improves the quality of visual representation through real-time evaluation. Specifically, the part navigator alleviates scale confusion issues and accurately locates class-discriminative regions, and the multi-part and multi-scale cross-attention models the contextual spatial relationship between global objects and semantic parts at different scales, for capturing subtle yet differentiated clues within fine-grained objects. Moreover, the multi-level semantic quality evaluation module is designed to supervise and enhance semantics from shallow to deep layers. And then quality probing classifiers are utilized to evaluate the feature quality output by the MPMSCA and MLSQE modules in an online manner for encouraging them to be more discriminative. More importantly, quantitative and qualitative results on four benchmark datasets validate the effectiveness and robustness of CSQA-Net.

In the near future, we will focus on several aspects to advance CSQA-Net: 1) Integrating the proposed MPMSCA module into a dual-branch vision Transformer architecture to perform multiple iterations of interactive learning between global semantics and local salient parts, thus facilitating FGVC. 2) Developing a part graph structure and utilizing its distinctive multi-hop connections to mine potential complementary information and relational embeddings among different discriminative parts.

\bibliographystyle{IEEEtran}
\bibliography{reference}

\vfill

\end{document}